\documentclass[letterpaper,10pt,conference]{ieeeconf} 

\IEEEoverridecommandlockouts                              
\usepackage[utf8]{inputenc}       
\usepackage[T1]{fontenc}          
\usepackage{cite}                
\usepackage{graphicx}            
\usepackage{amsmath, amssymb}      
\usepackage{algorithm}           
\usepackage{algpseudocode}       
\usepackage{xcolor}              
\usepackage{booktabs}            
\usepackage{caption}             
\usepackage[caption=false]{subfig} 
\usepackage{multirow}            
\usepackage{tabularx}            

\let\labelindent\relax 
\usepackage[inline]{enumitem} \usepackage{adjustbox}           
\usepackage{longtable}           
\usepackage{wrapfig}             
\usepackage{stfloats}            
\usepackage{hyperref}            
\hypersetup{
    colorlinks=true,
    linkcolor=blue,
    filecolor=magenta,
    urlcolor=cyan,
}
\usepackage{cleveref}            
\usepackage{mathptmx}            
\usepackage{tikz}
\usetikzlibrary{arrows.meta, positioning, shapes}
\usepackage{graphicx}
\usepackage{amsmath}
\usepackage{tabularx}

\graphicspath{{./images/}}

\title{\LARGE \bf
A Prescriptive Framework for Determining Optimal Days for Short-Term Traffic Counts}

\author{Arthur Mukwaya$^{1}$, Nancy Kasamala$^{1}$, Nana Kankam Gyimah$^{1}$, Judith Mwakalonge$^{1}$ ,\\ Gurcan Comert$^{2}$,  Saidi Siuhi$^{1}$, Denis Ruganuza$^{1}$, Mark Ngotonie$^{1}$ \\
$^{1}$ South Carolina State University, Orangeburg, South Carolina, US, 29117\\ 
$^{2}$ North Carolina A\&T State University, Greensboro, North Carolina, US, 27411\\  
}

\begin{document}

\maketitle
\thispagestyle{empty}
\pagestyle{empty}

\begin{abstract}
The Federal Highway Administration (FHWA) mandates that state Departments of Transportation (DOTs) collect reliable Annual Average Daily Traffic (AADT) data. However, many U.S. DOTs struggle to obtain accurate AADT, especially for unmonitored roads. While continuous count (CC) stations offer accurate traffic volume data, their implementation is expensive and difficult to deploy widely, compelling agencies to rely on short-duration traffic counts. This study proposes a machine learning framework, the first to our knowledge, to identify optimal representative days for conducting short count (SC) data collection to improve AADT prediction accuracy. Using 2022 and 2023 traffic volume data from the state of Texas, we compare two scenarios: an 'optimal day' approach that iteratively selects the most informative days for AADT estimation and a 'no optimal day' baseline reflecting current practice by most DOTs. To align with Texas DOT's traffic monitoring program, continuous count data were utilized to simulate the 24 hour short counts. The actual field short counts were used to enhance feature engineering through using a leave-one-out (LOO) technique to generate unbiased representative daily traffic features across similar road segments. Our proposed methodology outperforms the baseline across the top five days, with the best day (Day 186) achieving lower errors (RMSE: 7,871.15, MAE: 3,645.09, MAPE: 11.95\%) and higher R² (0.9756) than the baseline (RMSE: 11,185.00, MAE: 5,118.57, MAPE: 14.42\%, R²: 0.9499). This research offers DOTs an alternative to conventional short-duration count practices, improving AADT estimation, supporting Highway Performance Monitoring System compliance, and reducing the operational costs of statewide traffic data collection.

\textbf{Key words}- Short traffic count, Temporary counting, Annual average daily traffic, Machine learning, Unmonitored roads, Traffic networks
\end{abstract}


\section{Introduction}
\begin{figure}[h!]
    \centering
    \includegraphics[width=0.7\linewidth]{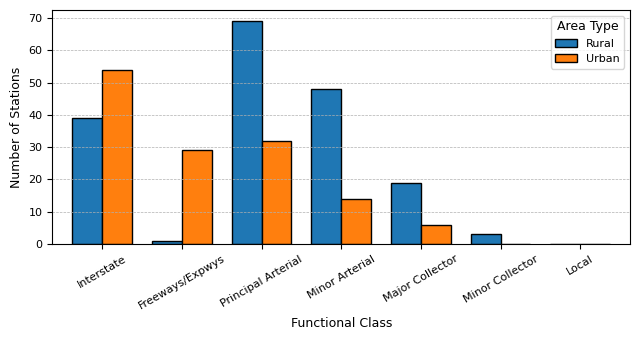}
    \caption{ Distribution of active permanent count stations in Texas (2023) by functional class and area type, highlighting the absence of coverage on local roads and limited representation in lower-priority urban areas.}
    \label{fig:station_distribution}
\end{figure}

In transportation planning, accurate and timely estimation of Annual Average Daily Traffic (AADT) is critical for effective roadway classification, safety analysis, and environmental impact assessments \cite{tsapakis2021informational}. State Departments of Transportation (DOTs) across the United States depend on AADT estimates to make critical decisions ranging from
highway expansion to funding allocation \cite{fhwa2022traffic}. Although permanent count stations provide continuous counts (CC) for direct AADT calculation, their deployment remains limited, leaving large portions of the roadway network unmonitored \cite{baffoe2023estimation}. In the state of Texas, approximately 350 active permanent count stations are in operation, with many situated along arterial roads. However, this is inadequate given the state's expansive 322,000-mile network, 64\% of which lies in rural areas, and with over 200,000 miles classified as local roads. As a result, large portions of both rural and low-priority urban roadways remain unmonitored \cite{fhwa2023,txdot2025}. Fig.~\ref{fig:station_distribution} illustrates the distribution of permanent stations across Texas by functional class and urban-rural classification.

To address this coverage gap, DOTs utilize short-term traffic counts, deploying portable equipment for one or two days of data collection. These observations are then adjusted using seasonal factors derived from nearby permanent stations, following guidelines in the Traffic Monitoring Guide (TMG) \cite{fhwa2022traffic}. The TMG mandates that National Highway System (NHS) and Principal Arterial System roads be counted on a 3-year cycle, with at least one-third surveyed annually. For lower functional classes such as minor arterials and major collectors, a 6-year cycle is recommended \cite{fhwa2018pocket}. Table~\ref{tab:state_count_schedules} shows selected states' current short count monitoring practices. These counts serve both regulatory and operational purposes, informing project level design and offering insight into broader traffic patterns. However, while cost-effective, this practice introduces significant AADT estimation errors, largely due to the ad hoc selection of count days without consideration of traffic variability or representativeness.

\begingroup 
\renewcommand{\arraystretch}{1.5} 
\begin{table*}[htbp]
\caption{Short traffic count monitoring practices by selected states}
\label{tab:state_count_schedules}
\centering
\footnotesize
\begin{tabularx}{\textwidth}{l X X X X}
\toprule
\textbf{State} & \textbf{Count Duration (hrs)} & \textbf{Cycle Frequency} & \textbf{Optimal Time/Season} & \textbf{Count Day Preference} \\
\midrule
Illinois & 24 or 48 & Marked: 2 yrs,\newline Unmarked: 4--5 yrs & late March to late October & Monday to noon Friday \\
Georgia & 48 & 1,2 or 4 year cycle based on segment need & Tuesday to Thursday,\newline typical conditions & Tuesday to Thursday \\
Minnesota & 48 ,\newline 24 if equipment fails & 2 year (Trunk Highways,\newline CSAH);\newline 4 year (Local);\newline 6 year (MSAS) & April to October & Monday to Wednesday \\
Florida & 24 (urban),\newline 48 (rural) & 3 year (on-system),\newline 6 year (off-system) & & Tuesday to Thursday \\
New York & 48 & Biennial or every 3 years (per HPMS) & & Tuesday to Thursday preferred \\
New Jersey & 48 & 3 year cycle for HPMS;\newline Annual for key routes & & Monday to Thursday preferred \\
Texas & 24 & & All year & Monday to Thursday \\
Alaska & 168 & 1--6 year rotation based on location & May to September & 7 consecutive days \\
Arkansas & 48 & & Annually & Monday to Thursday \\
Louisiana & 48 & 3 and 10 years & All year & Monday to Thursday \\
Michigan & 48 & 3 and 6 year & March to November & Monday to Thursday \\
New Hampshire & 48 & 3 years & April to November & Tuesday to Thursday \\
New Mexico & 48 & 3 year for NHS \& Principal Arterials,\newline 6 year for lower classification roads,\newline 12 year for local roads & & Monday to Friday \\
South Dakota & Interstate: 48,\newline Others: 24 & Interstate: 3 years;\newline State highways: half each year;\newline Locals: 6--8 years & April to October & Monday to Thursday \\
Washington & 48 & Every 3 years (all road types) & March to November & Tuesday to Thursday (manual counts);\newline Monday to Friday (mechanical counts) \\
\bottomrule
\end{tabularx}
\end{table*}
\endgroup 

While short-term traffic counting offers a cost-effective means of achieving broad geographic coverage, it often introduces significant AADT estimation errors. These inaccuracies primarily arise from DOTs selecting count days based on logistical convenience rather than data-driven assessments of traffic representativeness. This research targets the key limitations of current short-term counting practices, specifically:
\begin{enumerate}
    \item The selection of nonrepresentative count days undermines the accuracy of AADT estimates.
    \item Statewide short-count programs are resource-intensive and frequently yield low-value data when traffic conditions are nonrepresentative.
    \item Lack of standardized comparative performance metrics to quantify the advantages of optimized versus conventional counting strategies.
\end{enumerate}

Prior research has explored various approaches to traffic volume estimation, including statistical modeling \cite{mathew2021comparative,pulugurtha2021modeling}, deep learning \cite{han2024prediction}, and machine learning techniques \cite{das2020interpretable}. Other studies have applied geostatistical methods \cite{huynh2021estimating,baffoe2023estimating} or investigated the influence of external factors like weather and events on traffic volumes \cite{lin2022data}. However, none have examined the granular impact of count day selection on AADT accuracy. 

This study advances the field by introducing a data driven prescriptive framework that leverages machine learning to systematically identify optimal days for short-term traffic counts, an approach not previously attempted in the current research or adopted within current DOT practice.

The key contributions of this study are:
\begin{enumerate}
    \item Demonstrating improved AADT estimation accuracy by selecting an optimal day for short-term traffic counts.
    \item Developing a scalable machine learning framework to efficiently identify representative days based on predictive daily traffic patterns for AADT estimation.
     \item Establishing comprehensive performance quantification by comparing optimal versus baseline strategies using multiple metrics (RMSE, MAE, R², MAPE) across diverse road types.
\end{enumerate}

The rest of this paper is structured as follows: Section \ref{section:Related Works} reviews related work on traffic volume prediction and AADT estimation. Section \ref{section:Proposed Methodology} details the methodology, including data collection, preprocessing, feature engineering, and modeling. Section \ref{section:experimental setup} discusses the experimental setup, which highlights the machine learning model selection, target variable transformation, and performance metrics. Section \ref{section:Results and Discussion} presents the experimental results comparing optimized and baseline strategies. Finally, Section \ref{section:Conclusion and Future Work} concludes the paper and discusses future research directions.

\section{Related Works}
\label{section:Related Works}
This section reviews key contributions to AADT estimation, categorized into three methodological streams: regression-based models, deep learning approaches, and spatial modeling techniques.

\subsection{Regression and Ensemble Learning Models}

Sfyridis et al. \cite{sfyridis2020annual} demonstrated the effectiveness of Support Vector Regression (SVR) and Random Forest (RF) models for estimating AADT across the road networks of England and Wales. Their findings showed that SVR and RF significantly outperformed traditional linear regression, and clustering was used to reduce prediction errors. Raja et al. \cite{raja2018estimation} developed Ordinary Linear Regression (OLR) models (linear, quadratic, and logarithmic) to estimate Average Daily Traffic (ADT) on low-volume rural and local roads in Alabama. The linear and quadratic models performed comparably and outperformed the logarithmic model. 

Van der Drift et al. \cite{van2024global} employed Quantile Regression Forests (QRF) to generate a global time series of AADT on extra-urban roads, achieving a pseudo-$R^2$ of 0.7407 and offering prediction intervals for uncertainty quantification. Similarly, Yeboah et al. \cite{yeboah2023estimating} predicted traffic volumes on low-volume rural roads in Louisiana using both linear regression and RF models. Interestingly, linear regression outperformed RF, achieving an $R^2$ of 0.979 and an RMSE of 70.26, compared to 110.23 for RF. The study identified functional class, land use, and socioeconomic indicators as significant predictors.

\subsection{Deep Learning Models }
For AADT prediction and imputation, Han et al. \cite{han2024prediction} proposed a power function-based Long Short-Term Memory (LSTM) model that effectively handled missing data, achieving an MAE of 0.0733 and an error rate of 0.025\% even with 50\% missing values. Ramadan et al. \cite{ramadan2024seasonal} applied Artificial Neural Networks (ANNs) to forecast seasonal adjustment factors for AADT estimation, reporting higher accuracy than regression-based approaches. Mathew et al. \cite{mathew2023one} developed a One-Dimensional Convolutional Neural Network (1D-CNN) to estimate link-level AADT for locally classified roads in North Carolina. Their model outperformed eight other methods, including RF, Geographically Weighted Regression (GWR), and Ordinary Least Squares (OLS), with an RMSE of 712 compared to 729 for GWR and 765 for OLS.

\subsection{Spatial Models}

Huynh et al. \cite{huynh2021estimating} proposed a range of models—kriging, quantile regression, and point-based models—to estimate AADT in South Carolina at non-covered locations. Their hybrid kriging model, which defaults to a specified value when predictions exceed a threshold, improved RMSE by 21.37\% (RMSE = 217) over the baseline (RMSE = 276). Quantile regression achieved the best performance with a 23.19\% improvement (RMSE = 212), followed by the point-based (22.82\%, RMSE = 213) and regular regression models (17.03\%, RMSE = 229).

Zhen and Jidong \cite{zhen2024analyzing} utilized Graph Neural Networks (GNNs) to incorporate road network topology in AADT estimation across Atlanta's metropolitan region. Their diffusion-based graph convolutional model demonstrated superior generalizability across regions by learning transferable traffic flow patterns. Similarly, Baffoe et al. \cite{baffoe2023estimating} improved AADT prediction on low-volume roads using co-kriging with population data as a secondary variable.

Despite progress in AADT estimation, the specific problem of selecting optimal days for short-duration traffic counts remains underexplored. This study addresses that gap by leveraging machine learning to identify the most representative days for SC data collection on unmonitored roads.

\section{Proposed Methodology}
\label{section:Proposed Methodology}
\begin{figure*}[h!]
  \centering
  \includegraphics[width=\textwidth]{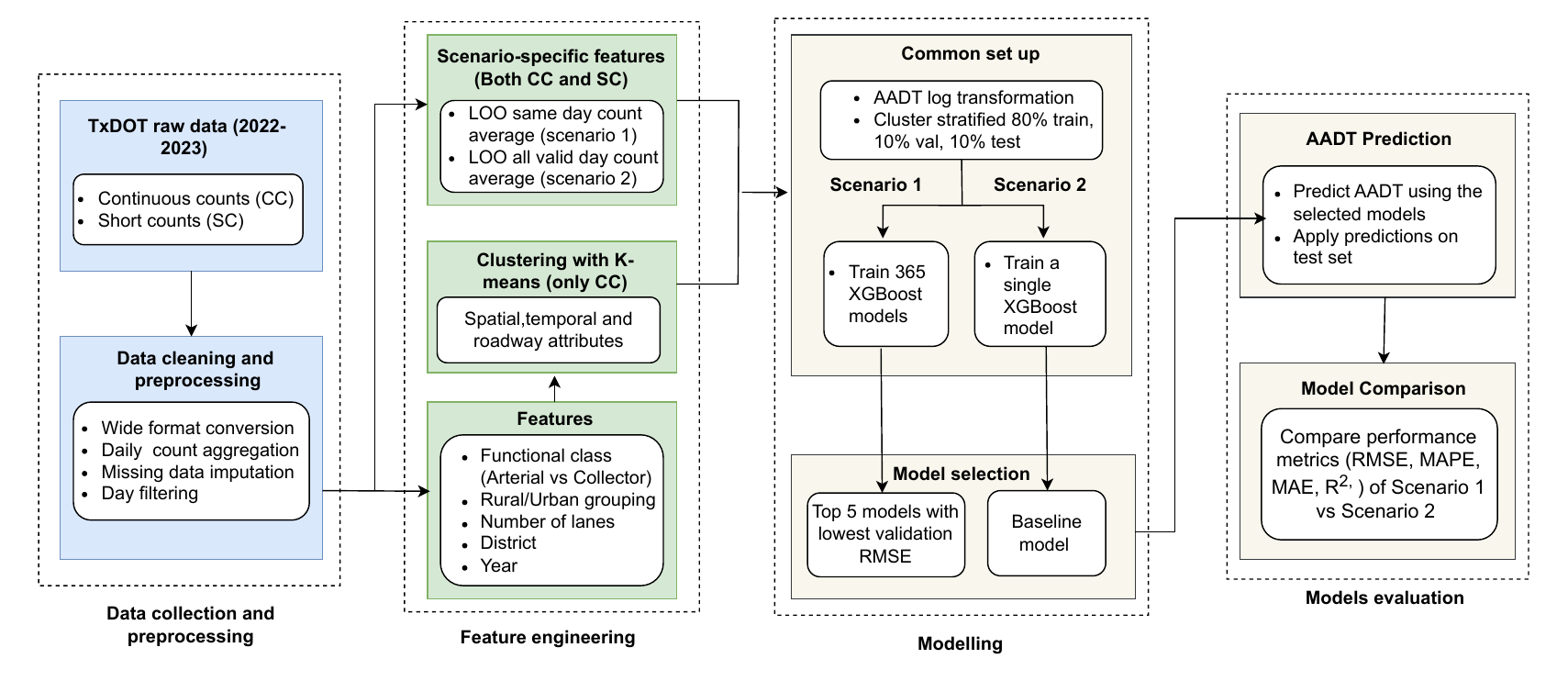}
  \caption{Proposed workflow for optimal short-duration traffic count day selection showing (1) Data preprocessing with wide format conversion, daily count aggregation,  missing value imputation and day filtering, (2) Feature engineering using roadway attributes, clustering, and leave-one-out averaging and (3) XGBoost modeling with performance evaluation using RMSE, MAE, MAPE, and R² metrics.}
  \label{fig:traffic_flow}
\end{figure*}

This study employs a data-driven approach to address the identified challenges and fulfill the stated contributions. As illustrated in Fig.~\ref{fig:traffic_flow}, the proposed workflow is organized into three sequential phases: (1) Data collection and preprocessing, where raw traffic data is acquired, cleaned, and prepared for modeling; (2) Feature engineering, which involves generating static and dynamic attributes to improve predictive performance; and (3) Experimental setup, where machine learning models are trained, validated, and evaluated using performance metrics.

\subsection{Data Collection and Processing}
This initial phase involved acquiring raw traffic count data, followed by cleaning and filtering processes to remove unreliable or incomplete records, as explained below.

\subsubsection{Study Area and Data Source}
\begin{figure}[h!]
    \centering
    \includegraphics[width=\linewidth]{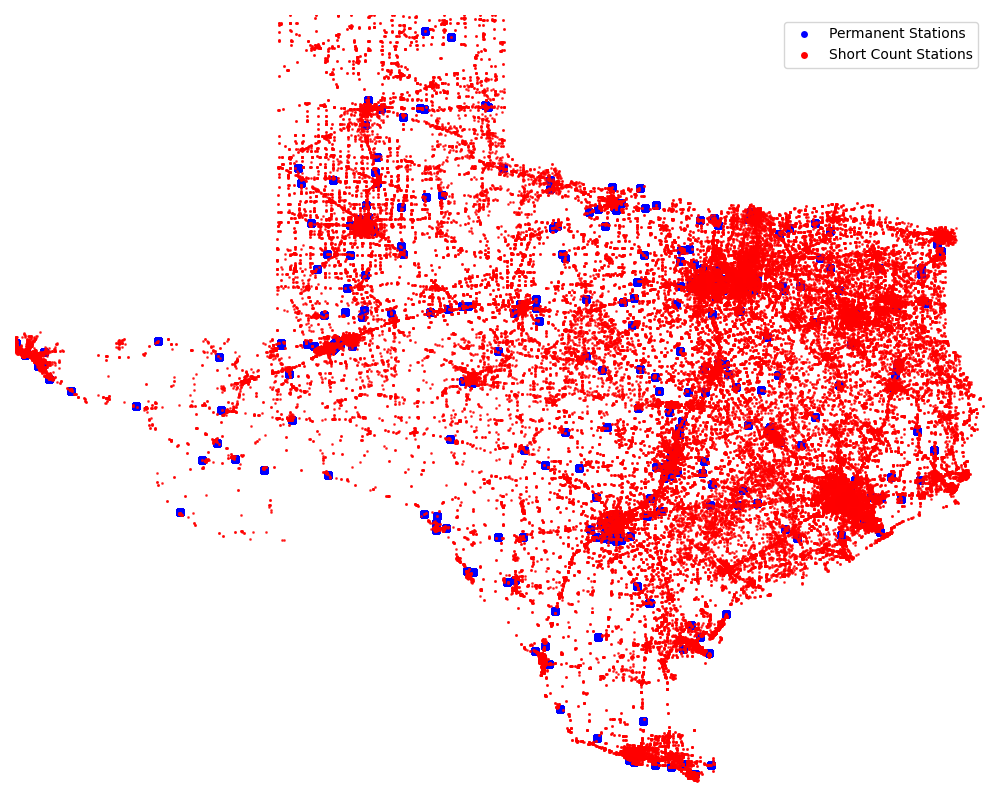}
    \caption{Study area of Texas State showing Permanent Stations  and Short Count Stations in 2022 and 2023.}
    \label{fig:station_distribution_texas}
\end{figure}

As shown in Fig.~\ref{fig:station_distribution_texas}, this study focuses on the road network within the State of Texas due to its extensive and diverse highway system that includes various functional class roads with a wide range of urban, suburban, and rural environments. The Texas Department of Transportation's (TxDOT) traffic data collection program provides access to a rich dataset of both continuous count (CC) and short-term count (SC). This extensive and varied dataset from a major U.S. state is ideal for developing and validating traffic estimation methodologies that are generalizable to challenges faced by other state DOTs across the nation.

The traffic volume data for this study was sourced from the TxDOT for the years 2022 and 2023 and comprises two main types of traffic information:

\begin{itemize}
    \item {Permanent Count Station Data:} TxDOT maintains an extensive network of over 350 permanent count sites providing continuous, year-round traffic counts that serve as the ground truth from which AADT values are derived. They use both intrusive sensors (inductive loop detectors) and non-intrusive sensors (HD radar) to collect this data \cite{txdot_permanent_counts}. The  installation, maintenance, and calibration of this equipment are done by both in-house TxDOT personnel and external contractors. The 1-day short counts used in this study were derived from this data. 
    
    \item {Short-Term Count Data:} TxDOT conducts 24-hour duration counts on diverse road segments from Monday through Thursday year-round, with days chosen largely at random for a given location and land use considerations selectively applied as shown in Fig. \ref{fig:SC_distribution}. The majority are performed by contract personnel utilizing pneumatic tube counters. All collected data undergoes screening and verification, and its Average Daily Traffic (ADT) is converted to AADT through a factoring process using monthly adjustment factors derived from CC Station data, where each location is assigned to a permanent count station or group.
\end{itemize}

\subsubsection{Dataset Description}
The dataset comprises two types of traffic volume data from TxDOT for 2022 and 2023. The permanent count dataset contains 631 unique permanent count stations, each providing 365 daily traffic counts from which an AADT is derived. The attributes associated with each permanent station include its precise geographic coordinates (latitude and longitude), functional class, daily counts, date, and area location.

\begin{figure}[h!]
    \centering
    \includegraphics[width=\linewidth]{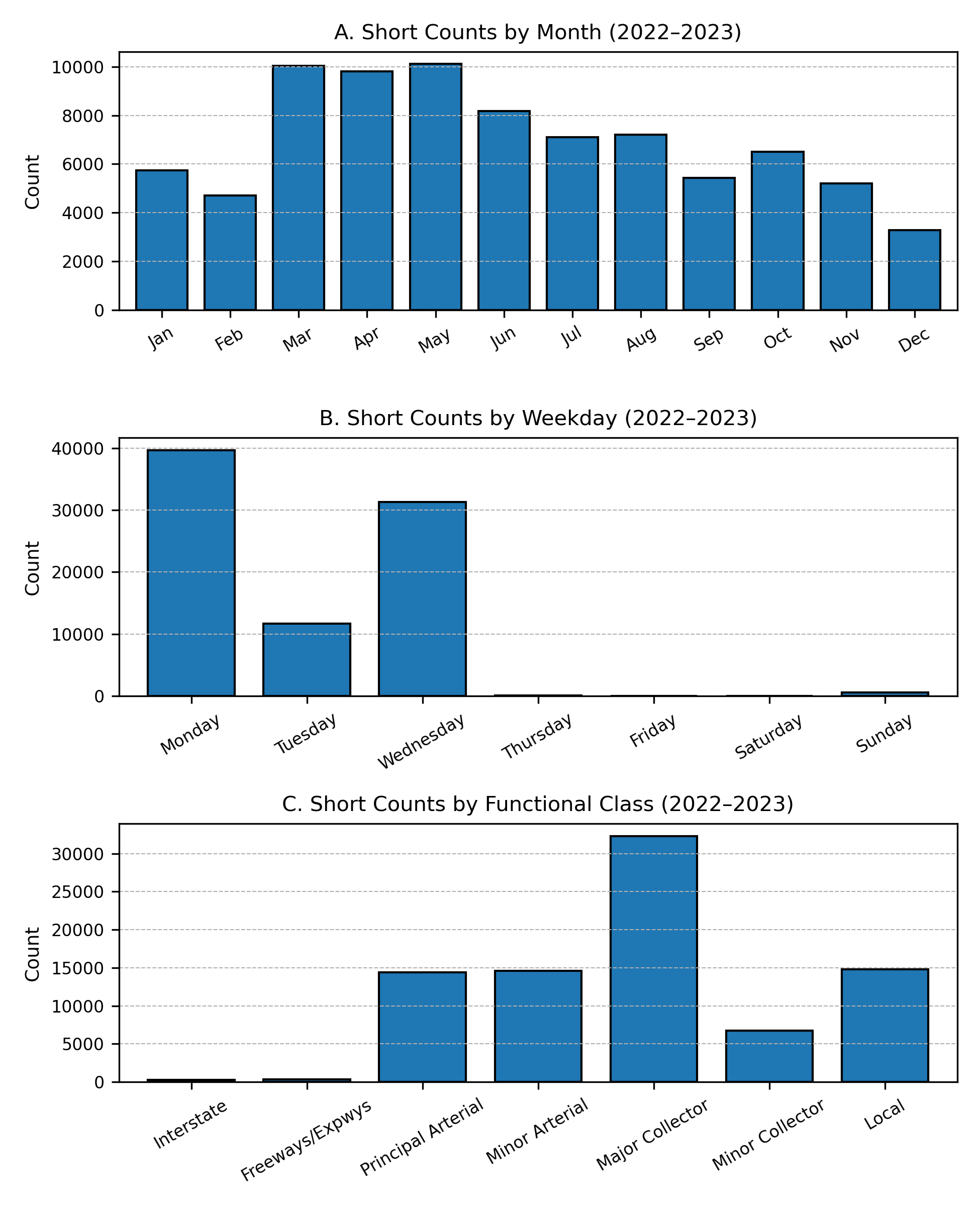}
    \caption{Distribution of short duration traffic volume counts in Texas (2022–2023) across (A) Months, (B) Weekdays, and (C) Functional class.}
    \label{fig:SC_distribution}
\end{figure}

The short count dataset consists of 54,040 individual short-term traffic count records, with each record detailing the observed traffic count, AADT, the date of observation, the functional class, and geographic coordinates. Fig. \ref{fig:SC_distribution} shows the distribution of SCs across months, days of the week, and functional classes. These counts are distributed across various locations and provide an intermittent view of traffic flow patterns. Both datasets also include attributes such as the number of lanes and rural/urban classification.

\subsubsection{Data Cleaning and Preprocessing}




We transformed the raw traffic data from long format (multiple rows per station) to wide format (one row per station with daily columns), enabling easier handling of missing daily data. Hourly counts were summed to derive daily totals.

Missing values in permanent station records were imputed using a supervised machine learning approach. Both Random Forest and XGBoost models were evaluated, with the model showing the best validation performance selected for final imputation.

The 24-hour count data for SC stations and their corresponding AADT values were merged based on a shared location ID. Records with incomplete data after merging were removed. To ensure accurate alignment, a unique station ID was created by combining latitude, longitude, and functional class. Year and day-of-year were extracted from volume count start dates for temporal consistency.

To ensure consistent modeling inputs, data from U.S. federal holidays, Fridays, Saturdays, and Sundays were excluded from both datasets, retaining only weekdays recommended for SC collection. This step ensures data reliability and mitigates common issues such as data leakage and misalignment, which are rarely addressed in prior AADT estimation frameworks.

\subsection{Feature Engineering}

This step generates relevant features for the machine learning models using static roadway attributes and dynamic traffic count characteristics. While both CC and SC data are utilized, only features derived from permanent station data are directly used as model inputs. SC data is leveraged to enhance feature engineering.

Once spatial and classification groupings were defined, we engineered two scenario-specific features to test the importance of day-level selection in AADT estimation. Leave-one-out (LOO) averages are used to prevent data leakage and provide robust contextual traffic information. Two distinct sets of features were developed: one capturing specific day-level effects (Scenario 1) and another representing general traffic trends across valid days (Scenario 2).

\subsubsection{Static Roadway Attributes}
To complement the attributes of each traffic count station, additional roadway static characteristics were extracted from the ArcGIS TxDOT Roadway Network API within a query radius of 25 meters. These variables included the number of lanes and rural/urban. These attributes were merged with the stations using latitude and longitude coordinates.

\paragraph{Functional Class Categorization}
Original functional classes were categorized into two main, broader, more balanced groups for subsequent similarity-based averaging:
\begin{itemize}
    \item \text{Arterial:} Includes principal arterial roads (interstate, other motorways/motorways, others) and minor arterial roads.
    \item \text{Collector:} Consists of major collector, minor collector, and local roads.
\end{itemize}

\paragraph{Rural/Urban Grouping}
Similarly, stations located in small urban, urban, and large urban areas were merged into a single urban category.

The combined functional classes and grouped rural/urban features are used for defining similar road segments, especially in leave-one-out for similar other day-specific averages, average other all valid similar days, and also as inputs for K-means clustering.

\subsubsection{K-Means  Clustering}
To uncover latent similarities among roadway segments, K-means clustering ($k=4$) was performed on the combined dataset of SC and CC stations. Clustering was based on standardized spatial and roadway attributes, including year, geographic location, combined functional class, lanes, and grouped rural/urban. The resulting clusters were used to stratify the training, validation, and test splits to ensure that each set retained a representative distribution of roadway types.

\subsubsection{Scenario-Specific Daily Count Features (LOO Averaging)}
We engineered two sets of features to evaluate the effect of targeted day selection in AADT estimation. These are both computed as leave-one-out (LOO) averages, ensuring no data leakage.

\paragraph{Scenario 1: Similar Other Day-Specific Averages}
This feature captures the average daily traffic on a specific day $X$ of the year (1–365) from all other stations sharing the same year, functional class, and rural/urban grouping. The formulation excludes the station $S$ being predicted:

\begin{multline} \label{eq:scenario1_avg_multline}
 \text{Similar Other Day Specific Averages} \\
 = \frac{\sum_{i \in \text{Group}, i \neq S} \text{Count}_{i,X}}{\text{N}_{\text{Group},X} - 1}
\end{multline}

\noindent where:

\begin{itemize}
    \item $\text{Group}$: Stations with matching Year, functional class, and rural/urban group.
    \item $S$: Target station.
    \item $X$: Day of the year.
    \item $\text{Count}_{i,X}$: Count at station $i$ on day $X$.
    \item $\text{N}_{\text{Group},X}$: Number of valid stations on day $X$.
\end{itemize}

\paragraph{Scenario 2: Average Other All Valid Similar Days}
This baseline feature captures the average of all valid days (non-weekend, non-holiday) across the same group:

\begin{multline} \label{eq:scenario2_avg_multline}
 \text{Average Other All Valid Similar Days} \\
 = \frac{\sum_{j \in \text{Valid Days}} \sum_{i \in \text{Group}, i \neq S} \text{Count}_{i,j}}{\text{Total Valid Counts in Group for Other Stations}}
\end{multline}

\noindent where:

\begin{itemize}
    \item $\text{Valid Days}$: Non-weekend, non-holiday days.
    \item $j$: A valid day of the year.
    \item All other terms are defined above.
\end{itemize}

Scenario 1 captures fine-grained temporal variability and tests the influence of specific count-day selection, while Scenario 2 serves as a baseline to evaluate the added value of targeted day selection.

\section{Experimental Setup}
\label{section:experimental setup}
This section outlines the machine learning pipeline employed in this study, including model selection, target variable transformation, data partitioning strategy, and performance evaluation criteria. All computational experiments were conducted using Google Colaboratory (Colab), a cloud-based Jupyter environment, leveraging its GPU and memory resources for efficient processing and model training.

\subsection{Machine Learning Model Selection}
XGBoost was selected as the primary algorithm for its robust gradient-boosted decision tree framework, which offers strong predictive performance, speed, and regularization capabilities suited to large and heterogeneous traffic datasets. It excels in capturing complex non-linear relationships, handling both categorical and continuous features, and mitigating overfitting through built-in regularization, making it highly suitable for AADT estimation across varied roadway conditions.

For a given traffic volume dataset $D = \{(x_p, V_p)\}_{p=1}^{n}$ with $n$ samples and $m$ predictors, where $x_p \in \mathbb{R}^m$ represents the feature vector and $V_p \in \mathbb{R}$ is the AADT, the objective function of a tree-based model can be defined as:

\begin{equation} 
    \label{eq:xgb_obj1}
    C = \underset{f}{\operatorname{argmin}} \sum_{p=1}^{n} l(V_p, \hat{V}_p^{(t)})
\end{equation}

where $V_p$ is the true AADT value and $\hat{V}_p^{(t)}$ is the prediction at iteration $t$. The prediction at iteration $t$ is updated as:
\begin{equation} 
    \label{eq:xgb_pred}
    \hat{V}_p^{(t)} = \hat{V}_p^{(t-1)} + f^{(t)}(x_p)
\end{equation}

where $f^{(t)}(x_p)$ is the output of the newly added tree at iteration $t$. To penalize model complexity and avoid overfitting, a regularization term $\Omega(f^{(t)}(x_p))$ is introduced, resulting in the regularized objective:
\begin{equation} 
    \label{eq:xgb_obj2}
    C = \underset{f}{\operatorname{argmin}} \sum_{p=1}^{n} l(V_p, \hat{V}_p^{(t-1)} + f^{(t)}(x_p)) + \Omega(f^{(t)}(x_p))
\end{equation}

XGBoost computes the optimal weights that minimize this regularized objective, and final predictions are generated after $t$ iterations.

Where:
\begin{itemize}
    \item $D$ is the traffic volume dataset.
    \item $n$ is the number of observations.
    \item $m$ is the number of features.
    \item $x_p \in \mathbb{R}^m$ is the feature vector for observation $p$.
    \item $V_p$ is the actual AADT value for observation $p$.
    \item $\hat{V}_p^{(t)}$ is the prediction at iteration $t$.
    \item $l(V_p, \hat{V}_p^{(t)})$ is the loss function (e.g., MSE).
    \item $\Omega(f^{(t)}(x_p))$ is the regularization term.
\end{itemize}

\subsection{Target Variable Transformation}
AADT values exhibited a positively skewed, non-normal distribution with a heavy upper tail. To improve model learning and ensure uniform error distribution across traffic volumes, the target variable was transformed using a natural logarithm plus one transformation: $\log(1 + y)$, where $y$ is the AADT value. This transformation compresses extreme values and reduces variance. After prediction, the inverse transformation $\exp(\hat{y}) - 1$ was applied to revert outputs to the original AADT scale.

\subsection{Data Splitting}
For each modeling scenario, the permanent count dataset was partitioned into training, validation, and test sets using an 80\%/10\%/10\% split. To ensure statistical representativeness and reduce sampling bias, a cluster-stratified splitting procedure was applied. The 10\% test set was drawn from 2023 data and stratified by K-means cluster labels. The remaining data was then split into training and validation sets using the same stratification. This approach maintains consistent distributions of roadway types and traffic characteristics across all subsets, enabling robust performance evaluation on previously unseen data.

\subsection{Performance Metrics}
Model performance was evaluated using standard regression metrics: Root Mean Squared Error (RMSE), Mean Absolute Error (MAE), Coefficient of Determination ($R^2$), and Mean Absolute Percentage Error (MAPE). These metrics are defined as follows:

\begin{itemize}
    \item \textbf{Root Mean Squared Error (RMSE)}:
    \begin{equation} \label{eq:rmse}
        \text{RMSE} = \sqrt{\frac{1}{N} \sum_{i=1}^{N} (y_i - \hat{y}_i)^2}
    \end{equation}
    It measures the square root of average squared prediction errors, sensitive to large deviations.

    \item \textbf{Mean Absolute Error (MAE)}:
    \begin{equation} \label{eq:mae}
        \text{MAE} = \frac{1}{N} \sum_{i=1}^{N} |y_i - \hat{y}_i|
    \end{equation}
    It provides a straightforward average of absolute prediction errors and is less influenced by outliers.

    \item \textbf{Coefficient of Determination ($R^2$)}:
    \begin{equation} \label{eq:r2}
        R^2 = 1 - \frac{\sum_{i=1}^{N} (y_i - \hat{y}_i)^2}{\sum_{i=1}^{N} (y_i - \bar{y})^2}
    \end{equation}
    It indicates the proportion of variance in AADT explained by the model; values closer to 1 indicate a stronger fit.

    \item \textbf{Mean Absolute Percentage Error (MAPE)}:
    \begin{equation} \label{eq:mape}
        \text{MAPE} = \frac{100\%}{N} \sum_{i=1}^{N} \left| \frac{y_i - \hat{y}_i}{y_i + \epsilon} \right|
    \end{equation}
    It measures average error as a percentage of the actual AADT. A small constant $\epsilon$ (e.g., $10^{-5}$) is added to avoid division by zero when $y_i = 0$.
\end{itemize}

Where:
\begin{itemize}
    \item $N$: Total number of observations.
    \item $y_i$: Actual AADT for observation $i$.
    \item $\hat{y}_i$: Predicted AADT.
    \item $\bar{y}$: Mean of all actual AADT values.
    \item $\epsilon$: Small constant for numerical stability.
\end{itemize}

\subsection{Model Training and Evaluation}
This phase implements and evaluates XGBoost-based AADT prediction models under both scenarios, using cluster-stratified train/validation/test splits.

\subsubsection{Scenario 1: Iterative Optimal Days Modeling}
For each of the 365 candidate days, a separate XGBoost model was trained using static roadway attributes and the engineered feature for that day (Scenario 1: similar to other-day specific averages). The target variable was log-transformed prior to training, and predictions were inverse-transformed. Only permanent count stations were used in model training; SC data contributed to feature engineering. Days yielding the lowest validation RMSE were identified as optimal representatives for short-term count scheduling.

\subsubsection{Scenario 2: Baseline Model Training}
A single XGBoost model was trained using static roadway attributes and the average feature across all valid similar days (Scenario 2). The same stratified data splits were used. As with Scenario 1, log-transformed AADT was used for training and inverse-transformed for final predictions. RMSE was computed on both validation and test sets to serve as the baseline for comparison.

\begin{figure}[h!]
    \centering
    \includegraphics[width=\linewidth]{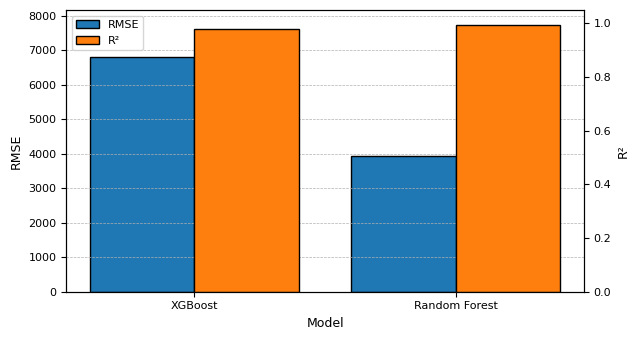}
    \caption{XGBoost and Random forest performance comparison for imputing missing traffic counts evaluated on a 10\% held-out sample.}
    \label{fig:filling}
\end{figure}

\begin{figure*}[h!]
    \centering
     \includegraphics[width=\textwidth]{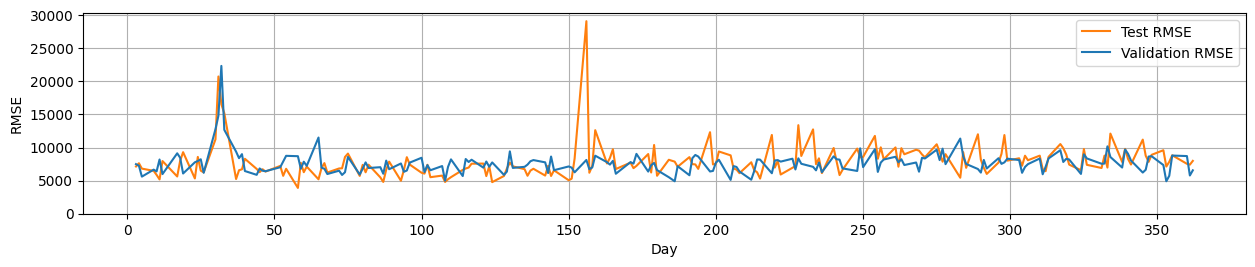}
     \caption{Daily variation of RMSE for AADT prediction across all 365 candidate days based on 1-day (24-hour) simulated short counts on test and validation sets.}
     \label{fig:rmse_r2_across_days}
\end{figure*}

\begin{figure*}[h!]
    \centering
    \includegraphics[width=\textwidth]{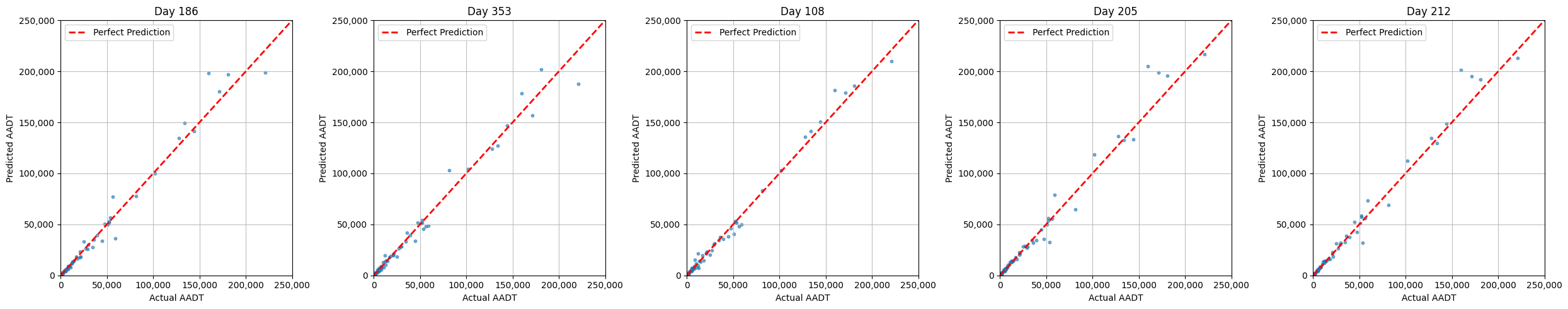}
    \caption{Actual vs. Predicted AADT for the Top 5 Optimal Days on the Test Set.}
    \label{fig:top5_actual_vs_predicted}
\end{figure*}

\begin{figure}[ht]
    \centering
    \includegraphics[width=0.45\textwidth]{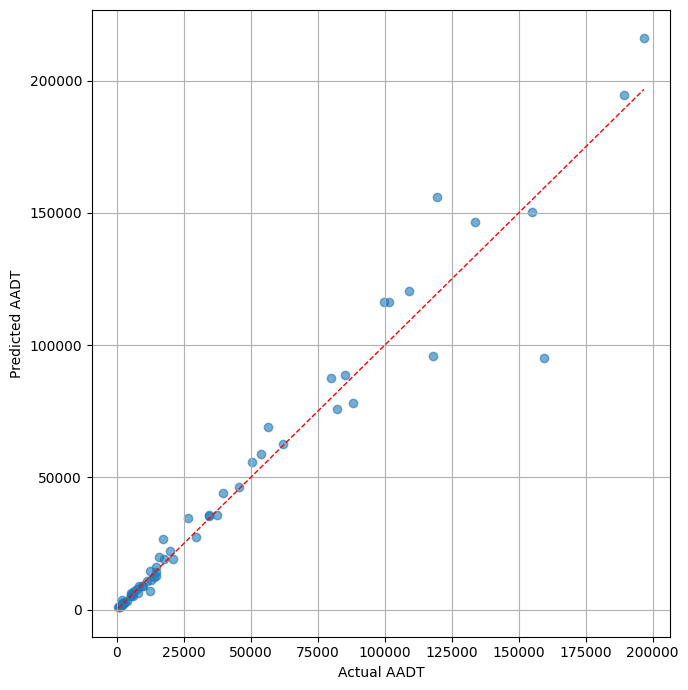}
    \caption{Actual vs. Predicted AADT for the Baseline model on the Test Set.}
    \label{fig:baseline_actual_vs_predicted}
\end{figure}


\section{Results and Discussion}
\label{section:Results and Discussion}
This section presents the results of the analyses conducted on the performance of the XGBoost models in estimating AADT for 24-hour SC traffic volume counts. A comprehensive comparison is conducted between the Optimal Days selection and the No-Optimum-Day baseline. The efficiency of the models is quantitatively assessed using standard regression metrics as detailed in Section \ref{section:Proposed Methodology} to demonstrate their predictive accuracy.

\subsection{Filling in Missing Count Data}
To address gaps in daily count records from CC stations, XGBoost and Random Forest were trained and tested on the same validation set, and their performance was compared using RMSE and R² metrics. A stratified sampling strategy was used to hold out 10\% of the data for evaluation. As shown in Fig: \ref{fig:filling}, the Random Forest model significantly outperformed XGBoost in terms of accuracy. XGBoost achieved an RMSE of 6809.58 and an R² of 0.980. Random Forest achieved a much lower RMSE of 3,933, with an R² of about 0.993. This strong performance from Random Forest made it the preferred choice for imputing the missing values.

\subsection{Scenario 1: Optimal Days}
    In Scenario 1, an iterative modeling process involved training a separate XGBoost model for each of the 365 days of the year. The evaluation on 10\% validation and 10\% test sets revealed distinguishable variability in AADT prediction performance across the different days. Fig. \ref{fig:rmse_r2_across_days} illustrates the daily variation of RMSE for all 365 candidate days, showing the fluctuations in prediction performance depending on the chosen count day.

Table \ref{tab:top5_scenario1} presents the top 5 best-performing optimal representative days based on the lowest validation RMSE with their respective validation metrics. The top-ranked day among these, Day 186, showed low RMSE values, indicating an outstanding fit and strong AADT predictive power.

 Figure \ref{fig:top5_actual_vs_predicted} shows the actual versus predicted AADT values for the top 5 performing days on their respective test sets. These scatter plots show a strong linear relationship, with most data points closely clustered around the perfect prediction line. This signifies the models' ability to accurately capture the true AADT across a wide range of traffic volumes from low-volume roads to high-volume segments.

\begin{table}[htbp]
    \centering
    \caption{Top 5 Best Performing Days by Validation RMSE}
    \label{tab:top5_scenario1}
    \footnotesize
    \begin{tabular}{lccccc}
        \hline
        \textbf{Metric } & \textbf{Day 186} & \textbf{Day 353} & \textbf{Day 108} & \textbf{Day 205} & \textbf{Day 212} \\
        \hline
        Validation & 4904.65 & 4906.74 & 4986.11 & 5112.31 & 5137.34 \\
        \hline
    \end{tabular}
\end{table}

\subsection{Scenario 2: No Optimal Days}

This scenario serves as a baseline representing a more traditional approach. AADT is estimated by averaging traffic counts across all valid similar days within a short duration counting period without any specific selection of days. The performance metrics for the baseline model show an RMSE of 11,185.00, MAE of 5,118.57, MAPE of 14.42\%, and R²of 0.9499 on the test set.

Figure \ref{fig:baseline_actual_vs_predicted} shows the actual versus predicted AADT values for the Scenario 2 baseline model on the test set, with more scatter around the perfect prediction line for both lower and higher AADT values.

\subsection{Comparison between Optimal and Non-optimal approaches}

This provides a quantitative comparison of the test set performance metrics obtained from Scenario 1 and Scenario 2. The results, as summarized in Table \ref{tab:comparative_metrics_detailed}, show the significant advantages of the Optimal Day approach over the No Optimum Day across the top 5 performing days.

\begin{table}[htbp]
    \centering
    \caption{Comparative Performance Metrics: Top 5 Optimal Days vs. Baseline (Test Set)}
    \label{tab:comparative_metrics_detailed}
    \resizebox{\columnwidth}{!}{
    \setlength{\tabcolsep}{3pt} 
    \renewcommand{\arraystretch}{1.1} 
    \begin{tabular}{lccccccc}
    \toprule
    \textbf{Metric} & \textbf{S2 Baseline} & \textbf{Day 186} & \textbf{Day 353} & \textbf{Day 108} & \textbf{Day 205} & \textbf{Day 212} \\
    \midrule
    RMSE & 11185.00 & 7871.15 & 7126.50 & 4804.06 & 8819.80 & 7776.85 \\
    \% Red. & -- & 29.63\% & 36.28\% & 57.05\% & 21.14\% & 30.59\% \\
    \midrule
    MAE & 5118.57 & 3645.09 & 3580.41 & 2850.75 & 3935.99 & 3580.31 \\
    \% Red. & -- & 28.79\% & 30.05\% & 44.32\% & 23.09\% & 30.05\% \\
    \midrule
    R$^2$ & 0.9499 & 0.9756 & 0.9800 & 0.9909 & 0.9694 & 0.9762 \\
    \% Inc. & -- & +2.71\% & +3.17\% & +4.32\% & +2.05\% & +2.77\% \\
    \midrule
    MAPE (\%) & 14.42 & 11.95 & 13.66 & 14.50 & 12.88 & 11.62 \\
    \% Red. & -- & 17.13\% & 5.27\% & -0.55\% & 10.75\% & 19.42\% \\
    \bottomrule
    \end{tabular}%
    }
\end{table}

The results summarized in Table \ref{tab:comparative_metrics_detailed} show the advantages of the Optimal Day approach over the No Optimum Day baseline, with the top 5 days having lower error metrics and higher explained variance across all evaluated performance metrics. 

\begin{enumerate}
    \item Day 186 model recorded a test RMSE of 7871.15 (29.63\% reduction), an MAE of 3645.09 (28.79\% reduction), and a R$^2$ of 0.9756 (2.71\% increase). Its MAPE of 11.95\% reflected a 17.13\% reduction.

    \item Day 353 model also demonstrated strong performance with a test RMSE of 7126.50, a 36.28\% reduction from the baseline. Its test MAE of 3580.41 represented a 30.05\% reduction, and its test R$^2$ of 0.9800 indicated a 3.17\% increase. The MAPE for Day 353 was 13.66\%, a 5.27\% reduction showcasing improvements across all metrics.

    \item Day 108 model achieved a test RMSE of 4804.06, representing a 57.05\% reduction compared to the baseline's RMSE. Its test MAE was 2850.75, reflecting a 44.32\% reduction from the baseline. Furthermore, Day 108 exhibited a Test R$^2$ of 0.9909, indicating a 4.32\% increase over the baseline's R$^2$. While its test MAPE of 14.50\% showed a minor 0.55\% increase relative to the baseline (14.42\%), this marginal difference is statistically and practically negligible considering MAPE's sensitivity to very low actual values. The improvements in RMSE, MAE, and R$^2$ overwhelmingly signify that AADT predictions on Day 108 are significantly more accurate than the baseline model.

    \item Day 212 model achieved a Test RMSE of 7776.85, a 30.59\% reduction, with an MAE of 3580.31 (30.05\% reduction) and an R$^2$ of 0.9762 (2.77\% increase). Day 212 also had a MAPE reduction of 19.42\% (11.62\% vs 14.42\%).

    \item Even Day 205 model, with an RMSE of 8819.80, which was the highest among the top 5 optimal days, still provided a 21.14\% reduction in RMSE, a 23.09\% reduction in MAE (3935.99), and a 2.05\% increase in R$^2$ (0.9694) compared to the baseline. Its MAPE was 12.88\% (10.75\% reduction).
\end{enumerate}

These detailed performance metrics demonstrate that the optimal day selection strategy leads to a significant reduction in prediction errors (RMSE and MAE) and a higher proportion of explained variance (R$^2$) across multiple top-performing days. While MAPE reductions varied, with Day 108 exhibiting a slight increase, the overall trend of improvements in absolute and explained variance metrics highlights the impact of strategic day selection on the accuracy of AADT estimations.

\section{Conclusion and Future Work}
\label{section:Conclusion and Future Work}
This research introduced a data-driven methodology to improve AADT estimation on unmonitored roads by selecting optimal short-duration count days with the lowest validation RMSE. Using Texas DOT traffic data from 2022–2023, we trained 365 XGBoost models to evaluate daily prediction performance. The best performing day on the validation set, Day 186, achieved a 29.65\% reduction in RMSE compared to the single model baseline, demonstrating the importance of strategic day selection over conventional approaches. 

The proposed methodology offers state DOTs a better option for short-term traffic practices by prioritizing data collection on the most representative days, as it improves prediction accuracy on uncovered rural and low-volume roads. Future research should focus on multi-day strategies and broader model validation across different U.S states. 

\section{Funding}
This research was partly funded by the U.S. Department of Education through the HBCU Master's Program Grant, by Grant No. P120A210048, the U.S. Department of Transportation's University Transportation Centers Program grant administered by the Transportation Program at South Carolina State University (SCSU) and NSF Grants Nos. 2131080, 2242812, 2200457, 2234920, 2305470.

\section{Acknowledgment}
The authors of this article would like to express their sincere appreciation to all the individuals who contributed to this research, including advisors, Texas DOT, colleagues, anonymous reviewers, and all other participants.

\bibliographystyle{IEEEtran}
\bibliography{sample}

\end{document}